# Measuring Psychological States Through Semantic Projection: A Theory-Driven Approach to Language-Based Assessment

*Maria Luongo*[1], *Davide Marocco*[1], *Nicola Milano*[1]

[1]University of Naples Federico II
Department of Humanistic Studies
Natural and Artificial Cognition Laboratory "Orazio Miglino"
via Porta di Massa 1, Naples, 80125, Italy

**Abstract**

Recent advances in natural language processing have enabled increasingly accurate estimation of psychological traits from language. However, most existing approaches rely on supervised models trained to predict questionnaire scores, limiting interpretability and generalizability across contexts. The present study introduces a theory-driven and fully unsupervised framework for measuring psychological states directly from natural language using semantic projection. Psychological constructs were operationalized as interpretable semantic axes derived from lexical anchors and items from validated clinical scales assessing depression, anxiety, and worry. Participants' textual responses (N = 247 observations from 145 participants across two time points) were embedded using Sentence-BERT and projected onto these axes to generate continuous psychological scores across multiple response formats, including selected words, generated words, phrases, and free-text responses. Projection scores were evaluated through correlations with standardized clinical measures (PHQ-9, CES-D, GAD-7, PSWQ), split-half reliability analyses, attenuation corrections, distributional similarity using Wasserstein distance, and comparisons with lexicon-based sentiment analysis (VADER). Results showed strong associations between projection scores and clinical measures, particularly for structured formats such as selected words, written words, and phrases (up to $r$ = .87 for depression and $r$ = .75 for worry). Free-text responses produced weaker results when analyzed as whole texts, but performance improved substantially when sentence-level aggregation strategies were applied. Reliability analyses revealed similar format-dependent patterns, with shorter structured responses showing the highest internal consistency. Distributional analyses further demonstrated that projection scores closely approximated the statistical properties of clinical measures, and projection-based methods outperformed lexicon-based sentiment approaches in longer text formats. These findings support semantic projection as an interpretable and scalable alternative to supervised language models for psychological assessment and highlight the importance of response format and text-processing strategies in language-based mental health measurement.

## Introduction

Human language is a primary medium through which individuals express their internal psychological states. Its flexibility allows people to describe thoughts and emotions in their own words, offering a potentially richer and more ecologically valid alternative to traditional self-report questionnaires based on predefined response formats (Kjell et al., 2022; Gu et al., 2025). The relevance of language for psychological measurement is well established. The lexical tradition in personality psychology posits that psychologically meaningful constructs are encoded in natural language (Goldberg, 1992), and recent work shows that semantic representations of questionnaire items can be used to model and generalize psychological constructs computationally (Abdurahman et al., 2024).

Recent advances in natural language processing (NLP) have substantially improved the analysis of textual data and the extraction of psychologically relevant information from language. In particular, embedding-based approaches represent linguistic units as vectors in a continuous semantic space, where meaning is encoded through patterns of co-occurrence and contextual usage. These developments, driven by transformer-based models (Bommasani et al., 2021; Vaswani et al., 2017) and sentence-level representations such as Sentence-BERT (Reimers & Gurevych, 2019), have enabled increasingly accurate modeling of psychological constructs, in some cases approaching the reliability of traditional rating scales (Kjell et al., 2022). Furthermore, recent work has shown that embeddings can capture relationships among test items and their intended constructs, supporting both the reconstruction of latent factor structures and the assessment of content validity (Milano et al., 2025; Milano et al., 2026). Although these models can capture complex patterns in language, it remains unclear how their outputs map onto underlying psychological constructs and which aspects of the input drive their predictions (Demszky et al., 2023).

Within the framework of distributional semantics, recent work has demonstrated that rich conceptual knowledge is explicitly encoded in the geometry of embedding spaces and can be extracted through simple geometric operations. In particular, Grand et al. (2022) introduced semantic projection as a domain-general method for recovering feature-specific knowledge by projecting word vectors onto directions defined by pairs of antonyms (e.g., small–large, safe–dangerous). These directions can be interpreted as "semantic scales," along which objects can be ordered according to specific features. Crucially, this approach does not rely on supervised learning but instead extracts knowledge directly from the structure of the embedding space, demonstrating that context-dependent semantic relationships are represented geometrically and can be accessed in an interpretable manner. Indeed, this perspective has important implications for psychological measurement. If semantic spaces encode structured knowledge that can be organized along interpretable dimensions, then such dimensions may provide a principled basis for representing not only perceptual or conceptual features, but also more abstract constructs, including psychological states. In this sense, language does not merely provide input for predictive models but constitutes the representational substrate within which constructs can be defined and measured. In this framework, measurement does not consist in approximating external scores, but in locating linguistic expressions within a structured semantic space. However, an important limitation of this framework concerns the nature of linguistic input. Semantic projection has primarily been applied to isolated words or well-defined categories, where meaning can be treated as relatively stable. In contrast, natural language use is inherently compositional and context-dependent. Meaning emerges from the interaction of multiple words within phrases and sentences, and varies depending on structure, emphasis, and discourse context. As a result, a single linguistic response may contain heterogeneous or even conflicting semantic signals (e.g., expressing both positive and negative affect within the same sentence). Such complexity cannot be reduced to a single lexical representation (Jang et al., 2025). This raises a critical challenge for extending projection-based approaches to psychological assessment. If psychological states are expressed through complex and contextually embedded language, then methods operating on isolated lexical units may fail to capture the full structure of meaning. At the same time, approaches that ignore contextual composition, such as lexicon-based sentiment methods (e.g., VADER or LIWC), are limited in their ability to account for these dynamics, as they assign fixed values to individual words independently of context (Hutto & Gilbert, 2014; Pennebaker et al., 2015). A key question, therefore, is how projection-based representations behave when applied to more complex linguistic inputs, in which meaning is distributed across multiple levels and may not be internally coherent. Recent studies have demonstrated that language-based scores can approximate clinical measures with high accuracy when combined with supervised learning approaches. For example, Gu et al. (2025) train predictive models on embeddings to estimate rating-scale scores across different response formats. While effective, these approaches conceptualize language-based assessment primarily as a predictive task, in which models are trained to reproduce external criteria. As a result, the derived measures remain inherently dependent on the availability, structure, and validity of the training data, potentially limiting their generalizability across populations, languages, and assessment contexts. The present study adopts a fundamentally different perspective. Rather than learning mappings from language to questionnaire scores, we operationalize psychological constructs directly within the semantic space using theory-driven axes. This shifts the goal from prediction to representation, enabling the derivation of interpretable scores without model training or labeled data. However, representing psychological constructs directly in semantic space also introduces new challenges. In particular, it remains unclear how variations in linguistic input, such as response length, structure, and level of contextual detail, affect the stability and psychometric properties of the resulting representations. While more elaborated responses may provide richer contextual information, they may also introduce variability that reduces measurement precision. Understanding this trade-off is critical for the development of reliable and valid language-based assessment methods. To address these issues, the present study had several aims: (1) to develop an unsupervised, theory-driven approach for quantifying psychological constructs from natural language, based on semantic projection along interpretable axes derived from psychologically meaningful word sets and clinical items; (2) to evaluate how the performance of projection scores changes across language formats that vary in contextual richness and linguistic complexity; (3) to examine the psychometric robustness of semantic projection scores through reliability and sensitivity analyses; (4) to assess the distributional similarity between projection scores and clinical measures; and (5) to compare semantic projection with a lexicon-based sentiment benchmark (VADER), in order to determine whether projection-based representations capture clinically relevant information beyond surface-level sentiment. Taken together, these aims position the present study as a step toward a theory-driven and fully unsupervised framework for language-based psychological measurement, in which constructs are not inferred through prediction but directly represented within a shared semantic space.

## Methods

The analysis proceeded in several stages. Semantic axes for depression and anxiety were constructed using lexical anchors and items from validated clinical scales (e.g., CES-D; Radloff, 1977; STAI-Y; Spielberger et al., 1983), and participants' textual responses were embedded and projected onto these axes to obtain continuous scores within a distributional semantic framework (Grand et al., 2022). These scores were evaluated through correlations with clinical measures of depression and anxiety (e.g., PHQ-9; Spitzer et al., 1999; GAD-7; Spitzer et al., 2006), alongside split-half reliability estimation and corrections for attenuation (Spearman, 1904; Nunnally & Bernstein, 1994). Distributional similarity was assessed using the Wasserstein distance (Givens & Shortt, 1984). Finally, projection-based scores were compared with lexicon-based sentiment estimates derived from VADER (Hutto & Gilbert, 2014).

### Dataset

The semantic projection procedure was applied to a pre-existing dataset collected via the online platform Prolific in January 2024, to evaluate the generalizability of language-based mental health assessment models on an independent sample (Gu et al., 2025). The sample included 145 participants (M_age = 41.6, SD = 12.6; range = 19–81), 101 females, 43 males, and one participant identifying as other, with the majority reporting the United Kingdom as their nationality. Participants were invited to complete a follow-up assessment two weeks later (Time 2), with 122 participants (84.1%) returning. Demographic characteristics remained comparable across time points (M_age = 42.3, SD = 12.6). Analyses were conducted on the combined dataset including both Time 1 and Time 2 observations (N = 247), while separate analyses for each time point are reported in the Supplementary Materials. Participants described their psychological states using natural language response formats designed to capture subjective experiences of depression and worry (Kjell et al., 2019). Participants were asked questions referring to the previous two weeks, for example: "Over the last two weeks, have you been depressed or not?"

Depending on the assigned response format, participants provided answers in one of several forms:

- select words (choosing words from a predefined list),
- write words (generating descriptive words),
- write phrases (short phrases describing their state),
- write texts (paragraph-length responses).

The same procedure was applied to assess worry/anxiety using a parallel prompt. Participants also completed validated clinical rating scales as external criteria. Depression severity was assessed using the Patient Health Questionnaire-9 (PHQ-9) (Spitzer et al., 1999) and the Center for Epidemiologic Studies Depression Scale (CES-D) (Radloff, 1977), while anxiety and worry were assessed using the Generalized Anxiety Disorder scale (GAD-7) (Spitzer et al., 2006) and the Penn State Worry Questionnaire (PSWQ) (Meyer et al., 1990).

### Construction of Psychological Semantic Axes

Two psychological constructs were modeled in the present study: depression-related affect and worry/anxiety-related affect. For each construct, a semantic axis was defined to represent the continuum between lower and higher levels of the corresponding psychological state. To examine the robustness of the approach, two alternative strategies were used to construct the semantic axes for each construct.

First, axes were derived from lexical anchors representing opposite emotional poles (e.g., "happy" vs. "sad", "calm" vs. "anxious") (Gu et al., 2026). Word embeddings were computed using the pretrained all-roberta-large-v1 model within the Sentence-BERT framework, which provides high-dimensional contextualized representations optimized for semantic similarity tasks (Reimers & Gurevych, 2019; Liu et al., 2019).

Second, axes were constructed using items from validated psychometric instruments. Items reflecting lower symptom levels were treated as positive anchors, whereas items reflecting higher symptom levels were treated as negative anchors. In this case, full questionnaire items (e.g., "I felt happy", "I felt depressed") were embedded as complete sentences, allowing the axes to capture richer contextual and phenomenological aspects of the constructs.

For each semantic axis, a direction vector was defined as the difference between the mean embedding of the positive anchor set and the mean embedding of the negative anchor set:

$$a = \frac{1}{m}\sum_{j=1}^{m} p_j - \frac{1}{n}\sum_{k=1}^{n} q_k$$

where $\mathrm{p}_j$ denotes the embedding of the $j$-th positive anchor, $\mathrm{q}_k$ denotes the embedding of the $k$-th negative anchor, $m$ is the number of positive anchors, and $n$ is the number of negative anchors. Thus, each semantic axis represents a theory-driven contrast between the positive and negative poles of the target construct (Grand et al., 2022).

Geometrically, this operation defines a straight line (i.e., a two-dimensional subspace) in the embedding space. The direction of this line captures the semantic transition from low to high levels of the construct (e.g., from calm to anxious, or from positive mood to depressive affect) (Osgood, 1952, 1964) (see Figure 1 for an example of axis construction).

Depression-related axes were derived from items of the Center for Epidemiologic Studies Depression Scale (CES-D; Radloff, 1977) and the Zung Self-Rating Depression Scale (SRDS; Zung, 1965), whereas anxiety-related axes were constructed using items from established anxiety measures, including the Zung Self-Rating Anxiety Scale (SRAS; Zung, 1971) and the State–Trait Anxiety Inventory (STAI-Y; Spielberger et al., 1983).

The clinical scales used to derive item-based axes and to evaluate the projection scores are described in detail below.

**Figure 1. PCA projection of semantic axes from worry-related single words.**

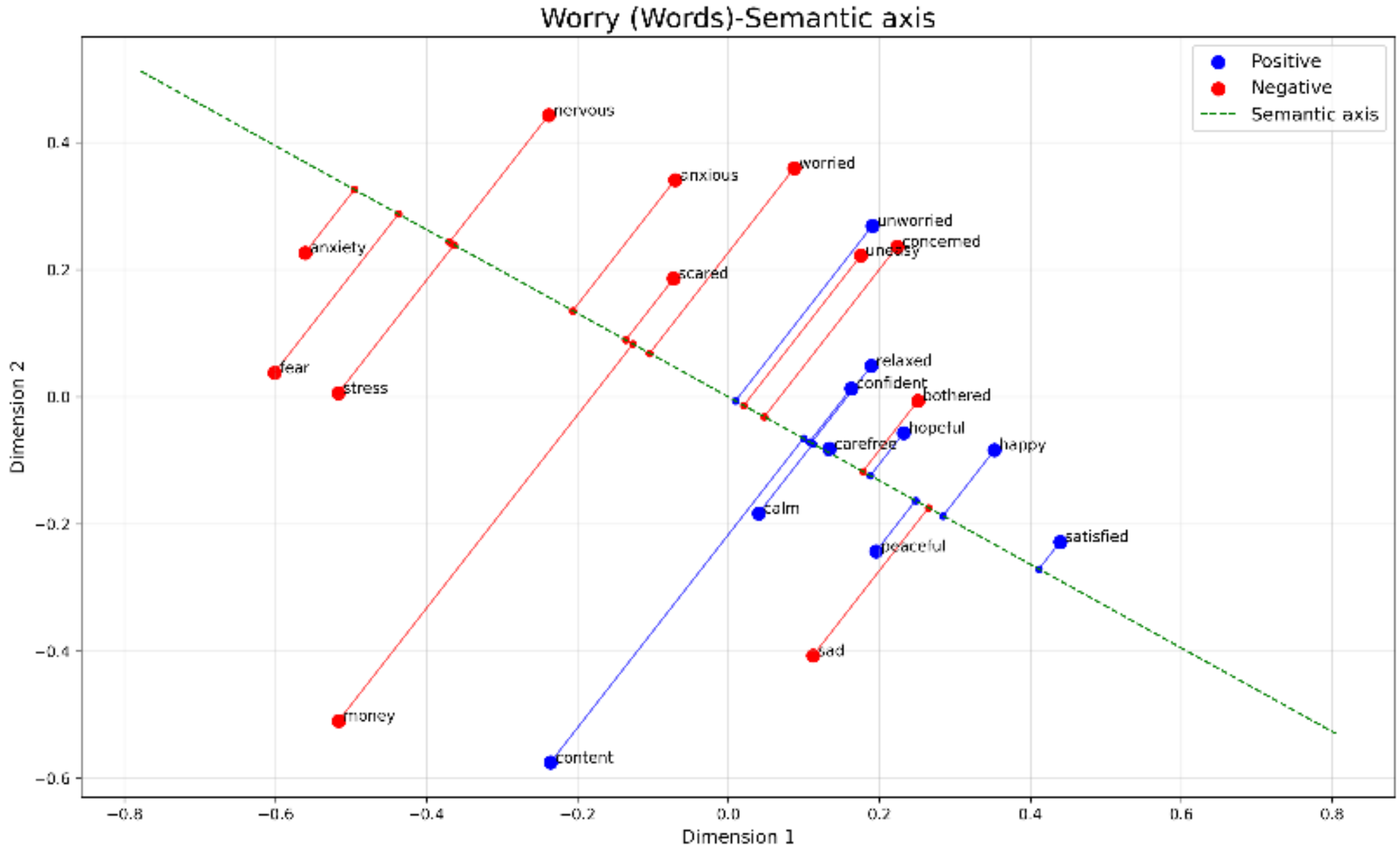

*Note.* The semantic axes are constructed from single-word lexicons representing worry-related constructs.

## Clinical Measures

Items from established psychometric instruments were used to construct item-based semantic axes representing the target psychological constructs. Specifically, items from the Center for Epidemiologic Studies Depression Scale (CES-D; Radloff, 1977) and the Zung Self-Rating Depression Scale (SRDS; Zung, 1965) were used to define depression-related dimensions, whereas items from the Zung Self-Rating Anxiety Scale (SRAS; Zung, 1971) and the State–Trait Anxiety Inventory (STAI-Y; Spielberger et al., 1983) were used to construct anxiety- and worry-related axes.

For each scale, items reflecting lower levels of symptomatology, including reverse-coded items, were treated as anchors for the positive pole of the construct, whereas items reflecting higher levels of symptoms were used as anchors for the negative pole. All items were embedded as complete sentences rather than isolated words, allowing the resulting semantic axes to capture richer contextual and phenomenological aspects of the constructs.

The CES-D is a 20-item self-report measure of depressive symptomatology in the general population (Radloff, 1977). The SRDS is a 20-item self-report scale designed to assess depressive symptoms, with both positively and negatively worded items (Zung, 1965). For anxiety-related dimensions, the SRAS was used as a 20-item measure of anxiety symptoms including both affective and somatic components (Zung, 1971), whereas the STAI-Y assesses anxiety as both a temporary state and a more stable trait (Spielberger et al., 1983).

## Computation of Semantic Projection Scores

To obtain a quantitative score for a given text (e.g., a participant's response), its embedding vector was projected onto the semantic axis using a dot product between the embedding x and the normalized axis:

$$\text{score}(x) = \frac{x \cdot a}{|a|}$$

where $\| a \|$is the Euclidean norm of the semantic axis. This projection can be interpreted as the coordinate of the text along the psychological continuum. Positive values indicate proximity to the positive pole (e.g., low depression), whereas negative values indicate proximity to the negative pole (e.g., high depression). For response formats consisting of a single unit (e.g., individual words or short phrases), projection scores were computed directly from the embedding of the entire response. For longer texts, projection scores were computed using three alternative approaches. First, a single score was obtained from the embedding of the entire text, capturing its overall semantic representation. Second, responses were segmented into sentence-level units $s_i$, and projection scores were computed for each sentence. These sentence-level scores were then summarized by computing their mean, capturing the overall semantic tendency of the text:

$$\text{score}_{mean}(T) = \frac{1}{k}\sum_{i=1}^{k} \text{score}(s_i)$$

where $k$ is the number of sentences in text $T$.

Finally, a maximum absolute sentence-level score was computed, capturing the most extreme semantic expression within the text:

$$score_{maxabs}(T) = score(s_j), \quad j = \arg\max_i |score(s_i)|$$

where $s_i$denotes the i-th textual unit, and $k$ is the total number of units within the text $T$.

All projection scores were computed within the same embedding space and using a fixed semantic axis, ensuring comparability across units and texts. These complementary representations allow the model to capture both the general semantic tendency of the text and its most salient or extreme emotional content.

Overall, this procedure yields a set of interpretable, theory-driven semantic dimensions that can be applied consistently across different types of textual input, enabling the quantification of psychologically meaningful variation in language.

## Psychometric Analyses

### *Correlation Analysis*

Pearson correlation coefficients were computed to examine the associations between semantic projection scores and standardized clinical measures of depression and anxiety across all semantic axes and response formats.

To account for attenuation due to measurement error in the clinical scales, partially disattenuated correlations were also computed. Specifically, observed Pearson correlations were corrected for the reliability of the clinical measures only, using the following formula:

$$r_{\text{partial}} = \frac{r_{\text{observed}}}{\sqrt{r_{\text{scale}}}}$$

where $r_{\text{observed}}$denotes the raw Pearson correlation and $r_{\text{scale}}$represents the reliability coefficient of the corresponding clinical measure.

This correction is grounded in classical test theory, which states that observed correlations are attenuated by measurement error (Spearman, 1904; Nunnally & Bernstein, 1994). As such, partially disattenuated correlations offer a more accurate approximation of convergent validity while avoiding potentially inflated estimates that may arise when correcting for uncertain reliability in computational measures (Schmidt & Hunter, 1996).

***Split-half reliability***

To estimate the internal reliability of the semantic projection scores, a split-half procedure was implemented separately for each semantic axis and response format. The aim was to quantify the internal consistency of projection-based indicators, that is, the extent to which different portions of the same response yield comparable projection scores when mapped onto the same semantic axis. This approach is consistent with recent psychometric applications in AI-based scoring, where reliability is estimated by partitioning inputs and correlating the resulting scores, typically with Spearman–Brown correction (Fan et al., 2023; Nunnally & Bernstein, 1994; Speer et al., 2025).

This procedure addresses a key challenge of NLP-derived measures: unlike traditional questionnaires, semantic projection scores are based on unstructured linguistic input that varies in length, content, and structure across individuals. As a result, reliability cannot be assumed a priori and must be empirically evaluated. The split-half approach provides a principled solution by treating different portions of the same response as parallel indicators of the same latent construct. If projection scores capture stable and meaningful semantic content, independent subsets of the same response should yield similar scores; conversely, low agreement would indicate a greater influence of noise or local variability.

To reduce potential biases related to order effects or narrative progression, an odd–even partitioning strategy was adopted. Textual units were alternately assigned to two halves (Half A: 1st, 3rd, 5th; Half B: 2nd, 4th, 6th, etc.), ensuring a balanced representation of the response.

Because response formats differed in structure, the procedure was adapted to the most appropriate unit type. For longer free-text formats (*Write text, Write text (Mean), Write text (Max Abs)*), responses were segmented into sentence-level units using sentence tokenization. For shorter formats, segmentation was performed at the level of lexical or phrase units. In *Write words* and *Select words*, responses were split into individual tokens using comma-based or whitespace tokenization. In *Write phrases*, segmentation relied on explicit delimiters (e.g., semicolons, line breaks) or, when absent, on rule-based heuristics such as internal capitalization.

After partitioning, projection scores were computed independently for each half using the same semantic axis and scoring procedure described above. Pearson's correlation coefficient between Half A and Half B scores was then computed across participants for each axis-by-format combination. Because this correlation reflects the reliability of half-length inputs, it was adjusted using the Spearman–Brown prophecy formula:

$$r_{SB} = \frac{2r_{half}}{1 + r_{half}}$$

where $r_{half}$ denotes the Pearson correlation between the two half-scores and $r_{SB}$ represents the estimated reliability of the split-half projection score. In line with standard practice, this correction was applied only when the half-test correlation was positive; otherwise, reliability was treated as undefined. The procedure provides a psychometric complement to validity analyses by quantifying the stability of semantic projection scores across independent portions of the same linguistic response (Furr, 2018; Speer et al., 2025).

***Sensitivity analysis***

Because projection scores showed variability in internal reliability across response formats, a sensitivity analysis was conducted to examine the extent to which associations with clinical measures were influenced by measurement error.

Split-half disattenuated correlations were computed by correcting for the reliability of both projection scores and clinical measures using the standard correction for attenuation formula:

$$r_{corrected} = \frac{r_{observed}}{\sqrt{r_{projection} \cdot r_{scale}}}$$

where $r_{projection}$ represents the split-half reliability of the projection scores and $r_{scale}$ represents the reliability of the clinical measure.

These corrected estimates provide an upper-bound approximation of the association under observed reliability conditions. Importantly, this analysis was conducted to assess robustness rather than to recover true population-level correlations (Schmidt & Hunter, 1996; Furr, 2018).

***Distributional similarity***

To assess the similarity between the distribution of AI-based projection scores and that of standardized clinical measures, we employed the Wasserstein distance (WD), which quantifies the minimum cost required to transform one probability distribution into another, providing a principled measure of distributional similarity (Givens & Shortt, 1984; Kolouri et al., 2016). Unlike correlation-based approaches, which capture linear associations between variables, the Wasserstein distance evaluates differences between entire distributions, taking into account both location and shape (Panaretos & Zemel, 2019). This makes it particularly suitable for comparing projection scores with clinical measures, as it allows for the assessment of whether the two distributions share similar statistical properties beyond their rank-order relationships. To ensure comparability across measures, all variables were standardized (z-scored) prior to computing the distance, yielding a normalized index $WD_z$. Lower values indicate greater similarity between the distributions of projection scores and clinical measures. The use of Wasserstein distance is well established in machine learning and computer vision, where it is widely applied to compare complex distributions and evaluate model outputs (Frogner et al., 2015; Arjovsky et al., 2017; Wang et al., 2013; Rabin et al., 2011; Ni et al., 2009).

**Lexicon-Based Sentiment Benchmark**

To provide a baseline for comparison with embedding-based semantic projection scores, a lexicon-based sentiment analysis was implemented using the VADER (Valence Aware Dictionary and

sEntiment Reasoner) model (Hutto & Gilbert, 2014). VADER is a rule-based approach that combines a predefined sentiment lexicon with a set of heuristics designed to account for linguistic features such as negation, intensifiers, punctuation, and capitalization. It is particularly well suited for short and informal text, where sentiment is often expressed through explicit lexical cues.

VADER was applied to all participant responses across the different response formats (i.e., *Select words*, *Write words*, *Write phrases*, *Write text*, *Write text (Mean)*, and *Write text (Max Abs)*). For each response, the model produced four standard sentiment indices: negative, neutral, positive, and a composite (compound) score. The compound score, which represents the normalized overall sentiment polarity ranging from −1 (most negative) to +1 (most positive), was used as the primary metric.

To align the sentiment measure with the assessment of psychological distress, the compound score was inverted, yielding a continuous distress index in which higher values reflect more negative affective content. The performance of VADER-based sentiment scores was evaluated through correlation analyses with clinical measures of depression and anxiety, using the same procedures adopted for semantic projection scores. Direct comparisons between VADER and embedding-based projections were also conducted by examining differences in their associations with clinical outcomes across response formats (see Results for full details).

**Results**

Following the methodology described above for axis construction, three reference axes were defined for depression: DEP_CESD, derived from CES-D items; DEP_WORDS, based on single-word representations; and DEP_ZUNG, derived from the Zung Self-Rating Depression Scale. Similarly, for the worry domain, three corresponding axes were defined: WOR_WORDS, based on single-word representations; WOR_ZUNG, derived from the Zung Self-Rating Anxiety Scale; and WOR_STAI, derived from the State–Trait Anxiety Inventory. These axes were subsequently used as reference dimensions for all analyses reported below.

**Correlation analysis**

As shown in Table 1, the semantic projections for depression were significantly associated with both CESDtot and PHQtot across nearly all response formats. The highest partially disattenuated correlations were observed in the *Select words* format (e.g., DEP_WORDS: $r = .87$ with CESDtot; $r = .81$ with PHQtot). High coefficients were also observed for the *Write phrases* and *Write words* formats (e.g., DEP_CESD: $r = .81$ with CESDtot; $r = .77$ with PHQtot). In contrast, the lowest coefficients were observed in the *Write text* format, particularly for the DEP_WORDS projection, which was not statistically significant (CESDtot: $r = .13$; PHQtot: $r = .12$). Intermediate but significant coefficients were observed for the aggregated text formats, including *Write text (Mean)* (e.g., DEP_CESD: $r = .79$ with CESDtot) and *Write text (Max Abs)* (e.g., DEP_CESD: $r = .74$ with CESDtot). Across formats, most projections showed significant associations with both depressive measures, indicating consistent convergence across scales. All remaining coefficients were statistically significant at $p < .001$.

As shown in Table 2, the semantic projections for anxiety and worry were significantly associated with both GADtot and PSWQtot across all response formats. The highest coefficients were observed in the *Select words* format (e.g., WOR_WORDS: $r = .72$ with GADtot; $r = .75$ with PSWQtot).

Comparably high values were also observed for the *Write phrases* and *Write words* formats (e.g., WOR_ZUNG: $r = .69$ with GADtot and PSWQtot). The lowest coefficients were observed in the *Write text* format, particularly for the WOR_WORDS projection ($r = .32$ with GADtot; $r = .36$ with PSWQtot), although these remained statistically significant. Aggregated text formats showed higher coefficients than the standard text format, including *Write text (Mean)* (e.g., WOR_WORDS: $r = .56$ with GADtot) and *Write text (Max Abs)* (e.g., WOR_WORDS: $r = .49$ with GADtot). All reported coefficients were statistically significant at $p < .001$.

**Table 1.** Associations between text-derived semantic projections and depressive symptom measures (CESD and PHQ) across response formats.

| | CESDtot | | | PHQtot | | |
|---|---|---|---|---|---|---|
| **Formats** | **DEP_CESD** | **DEP_WORDS** | **DEP_ZUNG** | **DEP_CESD** | **DEP_WORDS** | **DEP_ZUNG** |
| **Select words** | .84*** | .87*** | .82*** | .78*** | .81*** | .77*** |
| **Write phrases** | .81*** | .72*** | .78*** | .77*** | .69*** | .73*** |
| **Write words** | .81*** | .74*** | .75*** | .74*** | .69*** | .69*** |
| **Write text** | .59*** | .13 | .57*** | .59*** | .12 | .55*** |
| **Write text (Max Abs)** | .74*** | .39*** | .70*** | .72*** | .37*** | .68*** |
| **Write text (Mean)** | .79*** | .50*** | .73*** | .76*** | .48*** | .71*** |

*Note.* Values represent partially disattenuated Pearson correlations (corrected for the reliability of clinical scales only). Asterisks indicate significance levels ( * $p < .05$, **p < .01, ***$p < .001$).

**Table 2.** Associations between text-derived semantic projections and anxiety and worry measures (GAD and PSWQ) across response formats.

| | GADtot | | | PSWQtot | | |
|---|---|---|---|---|---|---|
| **Formats** | **WOR_STAI** | **WOR_WORDS** | **WOR_ZUNG** | **WOR_STAI** | **WOR_WORDS** | **WOR_ZUNG** |
| **Select words** | .70*** | .72*** | .72*** | .73*** | .75*** | .73*** |
| **Write phrases** | .59*** | .64*** | .69*** | .65*** | .67*** | .69*** |
| **Write words** | .61*** | .65*** | .69*** | .64*** | .65*** | .65*** |
| **Write text** | .52*** | .32*** | .61*** | .53*** | .36*** | .59*** |
| **Write text (Max Abs)** | .56*** | .49*** | .67*** | .61*** | .52*** | .65*** |
| **Write text (Mean)** | .61*** | .56*** | .67*** | .62*** | .57*** | .65*** |

*Note.* Values represent partially disattenuated Pearson correlations (corrected for the reliability of clinical scales only). Asterisks indicate significance levels ( * $p < .05$, **p < .01, ***$p < .001$).

**Internal reliability of semantic projection scores**

Split-half reliability estimates of the semantic projection scores are reported in Table 3. Reliability was highest in *Select words*, with coefficients reaching .89 for DEP_WORDS, .85 for WOR_WORDS, and .84 for both WOR_STAI and WOR_ZUNG. Similarly high values were observed for *Write phrases* and *Write words*, including .84 and .86 for DEP_CESD, .85 for DEP_WORDS, and .86 for WOR_WORDS. In contrast, the lowest reliability coefficients were observed in *Write text*, most notably for DEP_WORDS (.21), with other projections also showing reduced values (e.g., DEP_CESD = .55; DEP_ZUNG = .55; WOR_STAI = .48; WOR_WORDS = .51). *Write text (Mean)* and *Write text (Max Abs)* showed higher reliability than the standard text condition. In *Write text (Mean)*, coefficients reached .86 for WOR_ZUNG and .78 for WOR_STAI, while in *Write text (Max Abs)* they reached .83 for WOR_ZUNG and .77 for WOR_STAI.

**Table 3. Split-half reliability of semantic projection scores across response formats.**

| | DEP_CESD | DEP_WORDS | DEP_ZUNG | WOR_STAI | WOR_WORDS | WOR_ZUNG |
|---|---|---|---|---|---|---|
| **Select words** | 0,83 | 0,89 | 0,79 | 0,84 | 0,85 | 0,84 |
| **Write phrases** | 0,84 | 0,84 | 0,74 | 0,8 | 0,86 | 0,83 |
| **Write words** | 0,86 | 0,85 | 0,81 | 0,8 | 0,83 | 0,8 |
| **Write text** | 0,55 | 0,21 | 0,55 | 0,48 | 0,51 | 0,61 |
| **Write text (Mean)** | 0,74 | 0,49 | 0,68 | 0,78 | 0,77 | 0,86 |
| **Write text (Max Abs)** | 0,76 | 0,52 | 0,74 | 0,77 | 0,76 | 0,83 |

Note. Values represent split-half reliability coefficients (Spearman–Brown corrected), reflecting the internal consistency of projection scores across response formats and semantic axes.

**Sensitivity analysis**

The results of the sensitivity analysis are presented in Figure 2(A) and Figure 2(B). Across all response formats, semantic axes, and clinical measures, correlation coefficients increased systematically from raw Pearson correlations to partial correlations and to split-half corrected estimates. For depression-related projections (Figure 2(A)), the highest coefficients were consistently observed in the *Select words*, reaching values up to .93 (CESDtot, split-half). High values were also observed for *Write phrases* and *Write words*, with split-half estimates reaching up to .86 and .84, respectively. In contrast, the lowest values were consistently found in the *Write text*, where correlations remained substantially lower (e.g., around .61–.62 after split-half correction). Aggregated text formats showed intermediate performance, with *Write text (Mean)* reaching up to .84 and *Write text (Max Abs)* up to .74.

A similar pattern was observed for worry and anxiety measures (Figure 2(B)). Correlation coefficients again increased progressively across correlation types. The highest values were observed in *Select words*, reaching up to .80 (PSWQtot, split-half), followed by *Write phrases* and *Write words* (up to .74 and .72, respectively). The *Write text* again showed the lowest values, with split-half correlations remaining around .66 for GADtot and .67 for PSWQtot. *Write text (Mean)* reached values up to .69, whereas *Write text (Max Abs)* reached up to .67.

**Figure 2. Sensitivity analysis of semantic projection scores across response formats.**

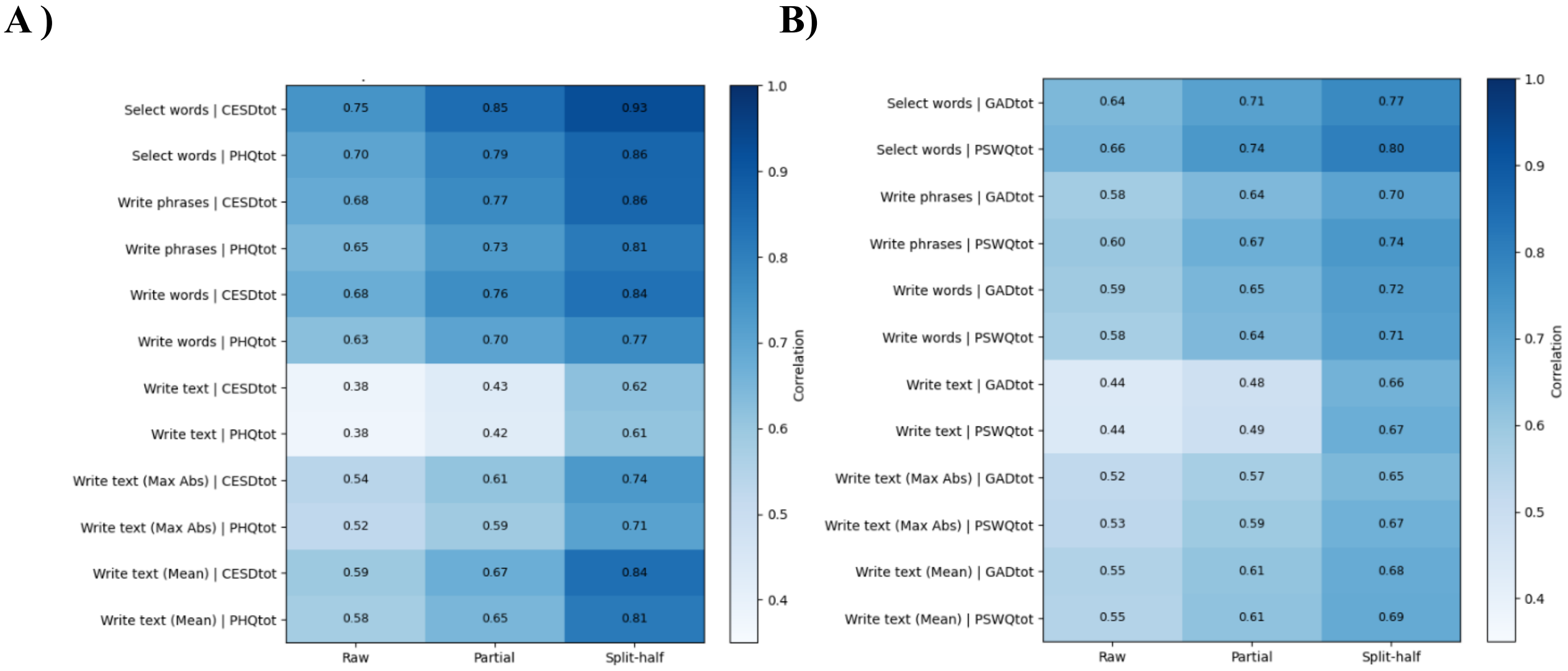

Note. (A) Depression-related projections. (B) Worry/anxiety-related projections. Values represent raw correlations, partial correlations and split-half corrected correlations.

## Distributional similarity between AI-based projections and clinical measures

Figure 3 shows the top-performing formats for each construct (DEP and WOR) based on Wasserstein distance ($WD_z$). Overall, projection scores closely matched the distributions of clinical measures in several formats.

For depression, the lowest $WD_z$ values were observed in *Write words* format ($WD_z$ = 0.12–0.16), alongside moderate-to-strong correlations (r = 0.61–0.66), indicating strong alignment in both association and distributional shape.

A similar pattern emerged for worry and anxiety. *Write words* and *Write phrases* formats showed low $WD_z$ values ($WD_z$ = 0.12–0.14) and moderate correlations (r = 0.53–0.62), reflecting close correspondence with clinical score distributions.

Overall, formats such as *Write words* and *Write phrases* not only correlated with clinical measures but also closely reproduced their distributional properties, supporting the validity of semantic projection scores at both relational and distributional levels.

**Figure 3. Distributional similarity of AI scores and clinical scores (top four formats per construct)**

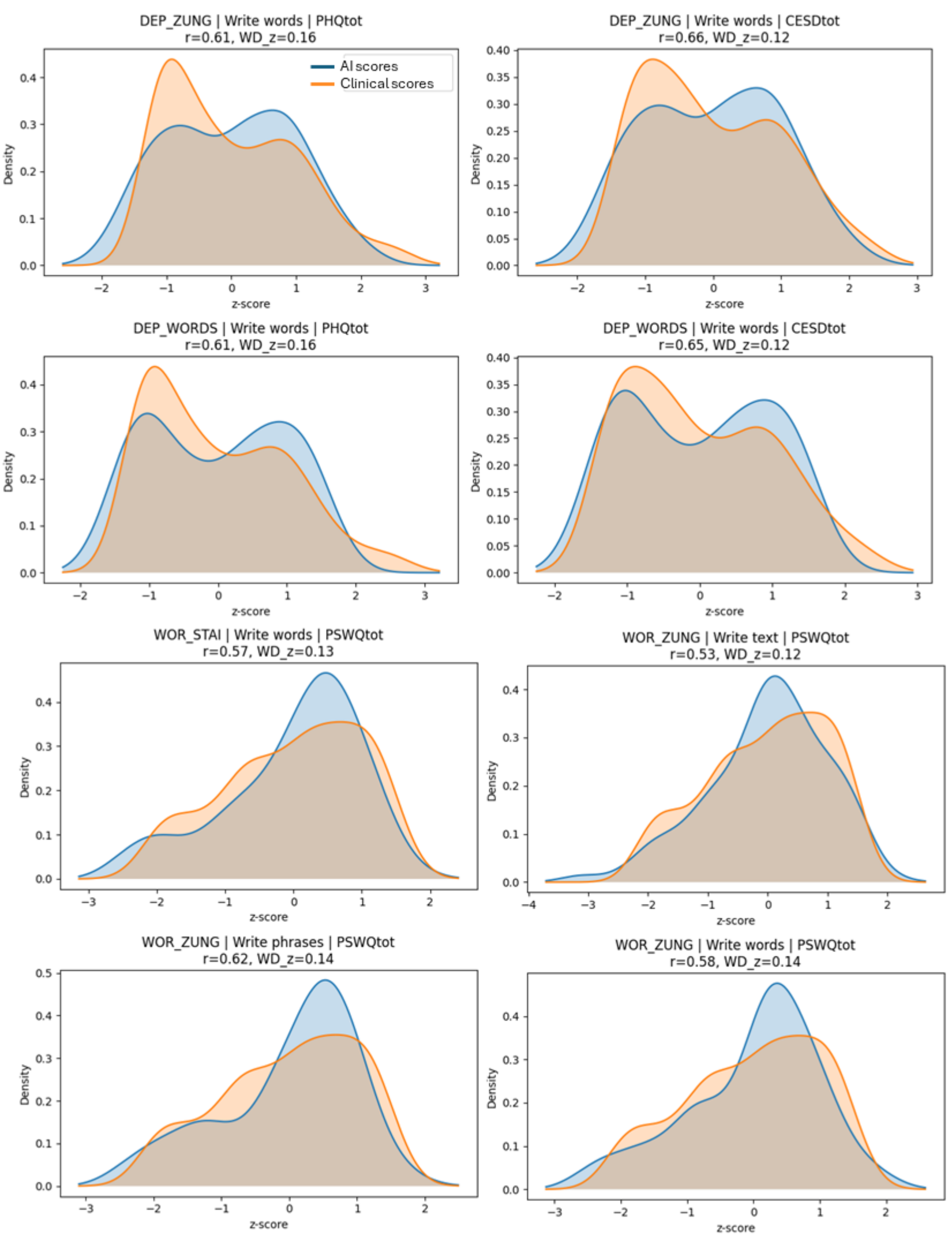

Note. Values represent standardized Wasserstein distances ($WD_z$). Reported correlation coefficients correspond to raw (non-disattenuated) Pearson correlations. The upper panel displays depression-related constructs (DEP), whereas the lower panel displays worry/anxiety-related constructs (WOR).

## VADER Vs Projection Scores

To compare the performance of VADER sentiment and embedding-based projection scores, we computed Δ as the difference between the partially disattenuated correlations obtained from projection-based scores and those derived from VADER sentiment. Figure 4 illustrates this comparison across formats. Projection scores were computed by selecting, for each format and scale, the highest partially disattenuated correlation among all available semantic axes within each construct. For *Write text*, projection scores were further optimized by selecting the best-performing representation across raw text, mean sentence-level aggregation (*Write text (Mean))*, and maximum absolute sentence-level aggregation *(Write text (Max Abs))*, while VADER scores were computed using the raw text only.

For *Select words*, *Write words* and *Write Phrases*, the performance of the two models is very similar, with minimal differences between projection and sentiment. In contrast, the largest differences emerge in *Write text*. In the depression domain, projection scores markedly outperform VADER, with $\Delta = 0.22$ for CESDtot and $\Delta = 0.28$ for PHQtot. A similar pattern is observed in the worry domain, where $\Delta = 0.25$ for GADtot and $\Delta = 0.24$ for PSWQtot.

**Figure 4. Δ between Embedding-Based projection and VADER sentiment across formats**

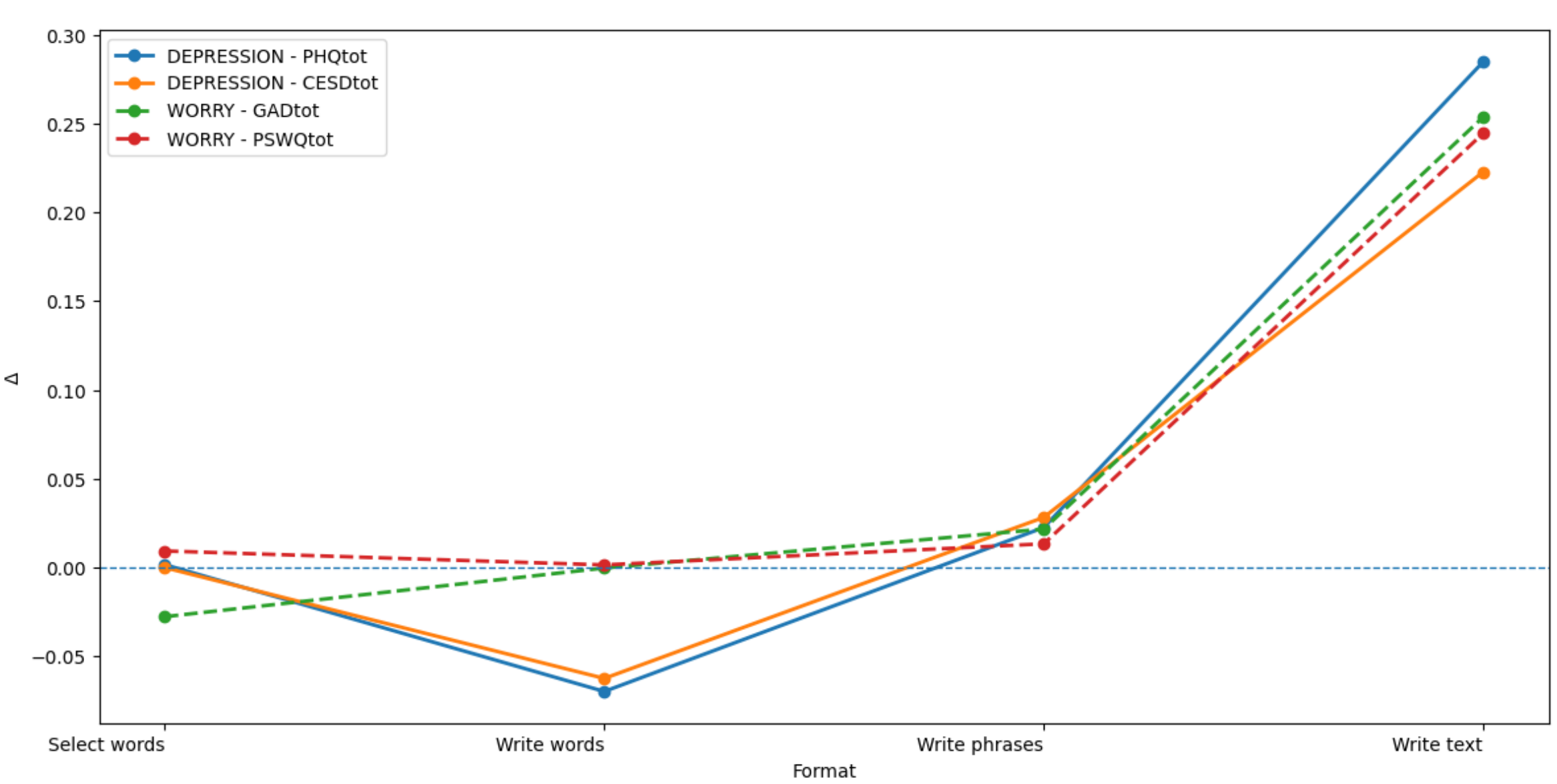

## Discussion

The present study evaluated whether theory-driven semantic projections derived from natural language can provide interpretable and psychometrically meaningful indicators of depression, anxiety, and worry across multiple responses formats. Unlike prior work that relies on supervised models to predict questionnaire scores from language (e.g., Gu et al., 2025), the present findings demonstrate that psychologically meaningful dimensions can be derived directly from semantic space using theory-driven projections (Grand et al., 2022). Overall, the results support the validity of this approach, while also highlighting important constraints related to the structure of linguistic input. First, the results of the correlation analyses indicate that projection scores show robust associations with established clinical measures of depression and anxiety. These associations were consistently strongest in some response formats, such as *Select words*, *Write words*, and *Write phrases*, and weaker in *Write text*. This pattern was further supported by the split-half reliability estimates. Indeed, the split-half results showed that internal consistency varied substantially across formats, with the highest reliability observed for any responses (e.g., *Select words*, *Write phrases*, *Write words, Write text (Mean)* and *Write text (Max Abs)*) and markedly lower values for *Write text* responses. These findings are consistent with prior work showing that open-ended formats can vary in reliability depending on response structure and elicitation method (Kjell et al., 2022; Kjell et al., 2024). The sensitivity analysis further clarified the implications of these reliability differences. As expected, correlations with clinical measures increased after correcting for attenuation, confirming that measurement error systematically reduces observed effect sizes (Spearman, 1904; Schmidt & Hunter, 1996). Formats that performed better in the raw analyses maintained stronger associations under both partial and split-half corrections, whereas *Write text* responses remained comparatively weaker. By contrast, *Write text (Mean)* and *Write text (Max Abs)* showed marked increases under split-half correction, reaching levels comparable to *Write words*. This indicates that differences across formats cannot be attributed solely to measurement error, but reflect systematic differences in the stability and informativeness of the semantic signal captured by each format. From a theoretical perspective, these findings highlight a key trade-off between contextual richness and measurement precision in language-based assessment. While free-text responses provide richer and more naturalistic representations of subjective experience, they also introduce greater semantic heterogeneity and noise, which can reduce both reliability and validity. Importantly, these results also underscore that, when working with longer and more complex responses (i.e., *Write text*), the choice of computational strategy becomes critical. Treating an entire free-text response as a single embedding and deriving one projection score may obscure meaningful internal variation and lead to a substantial loss of information. In contrast, approaches that operate at a finer-grained level, such as computing sentence-level projections and summarizing them via maximum absolute values or the mean (i.e., *Write text (Max Abs)* and *Write text (Mean)*), yield more reliable and informative representations. These strategies effectively reduce noise by capturing either the overall semantic tendency of the text or its most salient psychological content, thereby improving both the stability and interpretability of the resulting scores. A key contribution of the present study lies in the integration of distributional analyses. Using the Wasserstein distance, we showed that projection scores not only correlate with clinical measures but can also approximate their distributional properties. Formats such as *Write words* and *Write phrases* exhibited the lowest $WD_z$ values (= 0.12–0.16), indicating a close match in both central tendency and distributional shape. This finding extends traditional validation approaches by demonstrating that semantic projections capture not only rank-order relationships (i.e., correlations), but also the broader statistical structure of psychological constructs. In this sense, projection-based representations appear to approximate clinical measures at both relational and distributional levels. Crucially, the comparison with VADER sentiment analysis further highlights the specific advantage of embedding-based semantic approaches. While VADER sentiment and projection scores showed highly comparable performance in constrained formats such as *Select words*, *Write words*, and *Write phrases*, marked differences emerged for more complex and naturalistic language production. In *Write text*, projection-based scores consistently outperformed VADER sentiment, with Δ values ranging from 0.22 to 0.28 across both depression and worry measures. This suggests that lexicon-based methods, which rely on surface-level word valence, are limited in their ability to capture the complexity of extended language, whereas embedding-based projections benefit from contextualized representations that integrate semantic relationships across the text. Despite these strengths, several limitations should be acknowledged. First, the sample was relatively modest and drawn from an online population, which may limit generalizability across cultural and clinical contexts. Additionally, part of the data was collected at two time points from overlapping participants; however, analyses conducted separately yielded comparable patterns, supporting the robustness of the findings. In conclusion, the present results support the validity and flexibility of semantic projection as a framework for language-based psychological assessment. At the same time, they underscore that the effectiveness of such approaches depends critically on both the format of linguistic input and the methods used to process it. These insights have important implications for the design of future assessments, suggesting that combining structured elicitation formats with context-sensitive computational methods may offer the most effective path forward for measuring mental health through language.